\PassOptionsToPackage{bookmarks={false}}{hyperref}
\documentclass[conference]{IEEEtran}
\IEEEoverridecommandlockouts
\usepackage{cite}
\usepackage{amsmath,amssymb,amsfonts}
\usepackage{graphicx}
\usepackage{textcomp}
\usepackage{xcolor}
\usepackage{times}
\usepackage{soul}
\usepackage{algorithm,algpseudocode}

\usepackage[utf8]{inputenc}
\usepackage[small]{caption}
\usepackage{graphicx}
\usepackage{amsfonts}
\usepackage{color}
\def\BibTeX{{\rm B\kern-.05em{\sc i\kern-.025em b}\kern-.08em
    T\kern-.1667em\lower.7ex\hbox{E}\kern-.125emX}}
\begin{document}

\title{Deep Structured Cross-Modal Anomaly Detection}

\author{\IEEEauthorblockN{Yuening Li\dag, Ninghao Liu\dag, Jundong Li\ddag, Mengnan Du\dag, Xia Hu\dag}
\IEEEauthorblockA{\dag\textit{Department of Computer Science and Engineering, Texas A\&M University}, College Station, U.S.A \\
\ddag\textit{Department of Computer Science, Arizona State University}, Tempe, U.S.A \\
\{yueningl,nhliu43,dumengnan,xiahu\}@tamu.edu, jundongl@asu.edu}
}
\maketitle

\begin{abstract}

Anomaly detection is a fundamental problem in data mining field with many real-world applications. A vast majority of existing anomaly detection methods predominately focused on data collected from a single source. In real-world applications, instances often have multiple types of features, such as images (ID photos, finger prints) and texts (bank transaction histories, user online social media posts), resulting in the so-called multi-modal data.
In this paper, we focus on identifying anomalies whose patterns are disparate across different modalities, i.e., cross-modal anomalies. Some of the data instances within a multi-modal context are often not anomalous when they are viewed separately in each individual modality, but contains inconsistent patterns when multiple sources are jointly considered. The existence of multi-modal data in many real-world scenarios brings both opportunities and challenges to the canonical task of anomaly detection. On the one hand, in multi-modal data, information of different modalities may complement each other in improving the detection performance. On the other hand, complicated distributions across different modalities call for a principled framework to characterize their inherent and complex correlations, which is often difficult to capture with conventional linear models. To this end, we propose a novel deep structured anomaly detection framework to identify the cross-modal anomalies embedded in the data.  Experiments on real-world datasets demonstrate the effectiveness of the proposed framework comparing with the state-of-the-art.
\end{abstract}

\section{Introduction}
Over the past few decades, various successful anomaly detection methods have been developed and widely applied in a number of high-impact domains, including fraud detection, cybersecurity, algorithmic trading and medical treatment~\cite{chandola2009anomaly}. Identifying anomalies from a swarm of normal instances not only enables the detection and early alert of malicious behaviors but also helps enhance the performance of myriad downstream machine learning tasks.

Existing anomaly detection methods predominately focus on data from a single source. While in many real-world scenarios, the data we collected often comes from different sources or can be represented by different feature formats, forming the so-called multi-modal data. For instance, in synthetic ID detection, the target is to differentiate generated fake identities from the true identities of users~\cite{zhou2015muvir}. In this task, each identity is associated with information from different modalities, such as ID photos, bank transaction histories, and user online social behaviors. Compared with the data that is collected from a single source, when different modalities are presented together, they often reveal complementary insights into the underlying application domains.

In this paper, we focus on identifying anomalies whose patterns are disparate across different modalities and we refer the studied problem as \emph{cross-modal anomaly detection}, which is different from the conventional anomaly detection problem. The major reason of proposing such a problem is that a large portion of data instances within a multi-modal context are often not anomalous when they are viewed separately in each individual modality, but they present abnormal patterns or behaviors when multiple sources of information are jointly considered. For instance, in bank transaction records, each transaction is associated with different kinds of information such as user profiles, account summary, transaction records and the user signature, which are naturally of different modalities. In this case, if the user profile is not consistent with other sources of information of the same user, for instance, the ID is inconsistent with the signature, it might cause a bank fraud associated with a crucial crime. In other words, these anomalies present inconsistent behaviors across different modalities~\cite{gao2011spectral}, and the accurate detection of these anomalies has significant implications in practice. 

\begin{figure*}[t]
\centering
\includegraphics[width=1\linewidth]{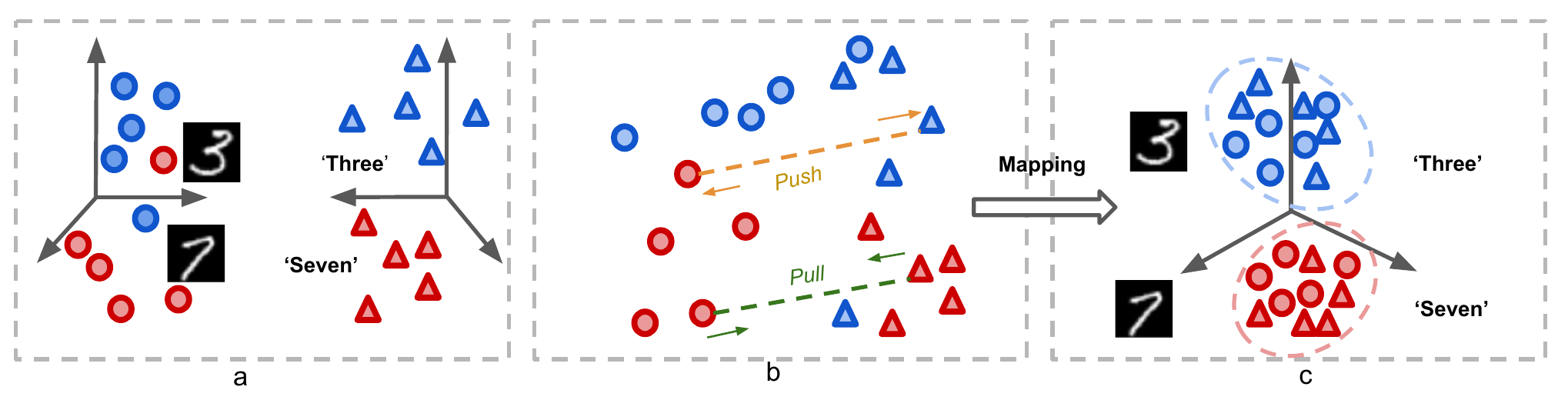}
\vspace{1mm}
\caption{An illustration of the proposed cross-modal anomaly detection framework - CMAD. In Figure 1(a), different colors represent different classes of instances (e.g., the blue denotes the class of number `three' while the red denotes the class of number `seven') and each shape denotes one particular data modality (e.g., the circle denotes the image modality and the triangle denotes the text tag modality). Figure 1(b) shows an example of processing inconsistent patterns (in orange dashed line) and consistent patterns (in green dashed line), where the goal is to find a consensus feature space in which consistent patterns across different modalities are pulled together while inconsistent patterns across different modalities are pushed away. Figure 1(c) shows the final data distribution in the consensus feature space, and we can find that the instances with consistent patterns across different modalities are indeed grouped together.}
\label{fig1}
\end{figure*}

However, cross-modal anomaly detection remains a challenging task. The main reason is that, to distinguish the cross-modal anomalies from other normal instances, we need to develop a principled framework to model the correlations between different modalities as the distributions of different modalities could be very complicated and may vary remarkably. Existing efforts either aim at maximizing the correlations between different modalities through linear mapping~\cite{thompson2000canonical,lai2000kernel}, or try to extract features from different modalities with low-rank approximation techniques~\cite{li2015multi,zhang2016collaborative}. In other words, the main focus of these methods is to embed various modalities into a consensus low-dimensional consensus feature space with a shallow learning model. In fact, the correlations might not be well captured in the low-dimensional consensus feature space by these shallow models due to their limited representation ability. In a nutshell, these shallow models cannot fully capture the nonlinear correlations among different modalities, which necessitates the investigation of the cross-modal anomaly detection problem with deep learning models.

In this paper, we aim to capture the nonlinear correlations among different modalities for cross-modal anomaly detection. In particular, we propose a deep structured cross-modal anomaly detection framework to identify inconsistent patterns or behaviors of instances across different modalities. The proposed deep structured cross-anomaly detection method consists of three major modules. Firstly, we train different deep neural networks to extract features from different modalities, and then project the features into a consensus latent feature space. Secondly, we pull samples with similar patterns across different modalities close to each other, while pushing samples with dissimilar patterns across different modalities far away from each other. Finally, we distinguish the cross-modality anomalies by measuring the similarity across different modalities. The main contributions are as follow:
\begin{itemize}
\item We systematically examine and define the cross-modal anomaly detection problem, and analyze the limitations of existing efforts.
\item We propose a novel cross-modal deep learning framework CMAD which could project data from different modalities into a comparable consensus feature space for anomaly detection when the original multi-modal data have different distributions.
\item We perform extensive experiments on various real-world multi-modal datasets to demonstrate the effectiveness of our proposed cross-modal anomaly detection framework.
\end{itemize}

\section{Related Work}    
In this section, we briefly review related work from two aspects: (1) conventional anomaly detection with data from a single modality; (2) cross-modal anomaly detection.

Anomaly detection refers to the problem of finding instances in a dataset that do not conform to the expected patterns or behaviors of the majority~\cite{chandola2009anomaly}. The anomaly detection problem has a wide range of applications in many high-impact domains, such as fraud detection~\cite{dorronsoro1997neural}, intrusion detection~\cite{buczak2016survey}, and healthcare monitoring and alert~\cite{pantelopoulos2010survey}. Conventional anomaly detection methods for a single modality data can be broadly categorized into classification based methods~\cite{chang2002anomaly,gonzalez2002combining} and clustering based methods~\cite{portnoy2001intrusion}. The classification based methods are often applied to learn a classifier from a set of labeled training data, and then classify the test data into normal classes or anomalous classes~\cite{chang2002anomaly,gonzalez2002combining}. The clustering based methods, on the other hand, often assume that anomalies belong to small and sparse clusters, while normal instances belong to large and dense clusters~\cite{portnoy2001intrusion,li2019specae}. While traditional classification based and clustering based methods are powerful in detecting anomalies among a swarm of normal instances, these methods lack the capability in modeling the cross-modal correlations among different modalities in multi-modal data for anomaly detection.

Identifying the cross-modal abnormal behaviors is a relatively new topic in the anomaly detection field. Only a few number of approaches have been developed to identify the inconsistent patterns across different modalities. Some of the recent works detect cross-modal anomalies by analyzing the clustering results in different modalities. These methods are based on the assumption that the underlying clustering structure of normal instances is usually shared across multiple modalities, while the anomalous instances belong to different clusters. In particular, in~\cite{gao2011spectral}, a multi-view anomaly detection method is proposed. It first obtains the spectral embeddings of instances with an ensemble similarity matrix, and then calculates the anomaly score of each instance based on the cosine distance between different embeddings. Later on, Alvarez et al. proposed an affinity propagation based anomaly detection algorithm, which could identify anomalies by analyzing the neighborhoods of each instance in different views~\cite{marcos2013clustering}. Another widely used cross-modal anomaly detection method is based on the low-rank learning, these methods represent data across different modalities in a low-rank feature space, and typical methods along this line include~\cite{zhang2016collaborative} and~\cite{li2015multi}. Both the aforementioned clustering based methods and low-rank learning based methods aim to identify inconsistent patterns across different modalities by learning meaningful feature representations. As the linear CCA based methods, these methods are often limited by their representation ability due to the linear projections.

\section{The Proposed Framework}
In this section, we introduce the proposed cross-modal anomaly detection framework CMAD in details. We begin with a formal definition of the cross-modal anomaly detection problem, and then propose a principled method to learn the consensus data representations across different modalities. After that, we discuss how to leverage deep structure to extract effective feature representations. Finally, we will introduce how to perform cross-modal anomaly detection with the built deep learning model.
\subsection{Problem Definition}
Given $K$ different modalities, $\mathbf{M}_{1}$, $\mathbf{M}_{2}$,..., $\mathbf{M}_{K}$ are the corresponding feature matrices of these $K$ modalities in $C$ different classes, where $\mathbf{M}_{k} \in \mathbb{R}^{N \times d_{k} }$, and $d_k$ is the feature dimensionality of the $k^{th}$ modality. $N$ is the number of instances in the dataset. The feature representation of the $i^{th}$ instance is denoted as $\mathbf{M}^{i}$, where $i$ $\in$ \{1,2,...,$N$\}. $\mathbf{\mathbf{M}}_{k}^{i}$ denotes the feature vector of the $i^{th}$ instance from modality $\mathbf{M}_{k}$. The class label of the $i^{th}$ instance in modality $\mathbf{M}_{k}$ is denoted as $\mathbf{y}_{k}^{i}$, where $\mathbf{y}_{k}^{i}$ $\in$ \{1,2,...,$C$\}. Our task is to find the anomalous instances whose patterns (w.r.t. class labels) are inconsistent across different modalities. For the notation convenience and without loss of generality, we only include two modalities $\mathbf{M}_{A}$ and $\mathbf{M}_{B}$ when introducing the methodology, where $A$, $B$ $\in$ \{1,2,...,$K$\} and $A$ $\neq$ $B$, but our method can be easily extended to consider any number of modalities.

Based on the aforementioned terminologies, we define the problem of \emph{cross-modal anomaly detection} as follows:

\textbf{Definition: (Cross-Modal Anomaly Detection)} Given a dataset $\mathcal{X}$ which contains multiple modalities as data sources, such as $\mathbf{M}_{A}$, $\mathbf{M}_{B}$, we define that the $i^{th}$ instance in $\mathcal{X}$ is regarded as an anomaly when: 
\begin{equation}
 F(\mathbf{\mathbf{M}}_{A}^{i} , \mathbf{\mathbf{M}}_{B}^{i})  < \epsilon,
\end{equation}
where $F(.,.)$ is a function measuring the cross-modal similarity between a pair of observations, whereas in this definition, the observations belong to the same instance. The parameter $\epsilon$ is a pre-defined threshold value such that if the measured similarity is larger than or equals to $\epsilon$, we regard that the $i^{th}$ instance is normal across the two modalities; otherwise, the $i^{th}$ instance is a cross-modal anomaly.

\subsection{Data Fusion across Different Modalities}
To detect the cross-modal anomalies, the fundamental prerequisite is to develop a principled way to fuse the information from different modalities. To achieve this target, a prevalent way is to learn the mapping function which could project instances from different modalities into a consensus feature space. In this way, it presents us a unified feature space in which we can measure the cross-modal similarity of the data instances, based on which the cross-modal anomaly detection can be easily carried out.
 
To this end, we transform instances from two modalities $A$ and $B$ into a consensus feature space with two linear transformation matrices, denoted as $\mathbf{U}$ $\in$ $\mathbb{R}^{d_{A}\times r}$ and $\mathbf{V}$ $\in$ $\mathbb{R}^{d_{B}\times r}$, where $r$ is the dimensionality of the resultant unified feature space and its value is much smaller than $d_{A}$ and $d_{B}$. After transformation, we obtain the following data representations for modalities $A$ and $B$:
\begin{equation}
\mathbf{M}_{\tilde{A}}=\mathbf{U}^{T}\mathbf{M}_{A}\label{eq1},
\end{equation}
\begin{equation}
\mathbf{M}_{\tilde{B}}=\mathbf{V}^{T}\mathbf{M}_{B}\label{eq2}.
\end{equation}

\noindent Given a pair of instances $i$ and $j$, we employ the cosine similarity to measure their cross-modal similarity in the transformed feature space:
\begin{equation}
\begin{aligned}
F({\mathbf{M}}_{A}^{i} , {\mathbf{M}}_{B}^{j}) &= \mathrm{cos}({\mathbf{M}}_{\tilde{A}}^{i}, {\mathbf{M}}_{\tilde{B}}^{j}) \\
&= \frac {\mathbf{M}_{\tilde{A}}^{i} \cdot \mathbf{M}_{\tilde{B}}^{j}}{||\mathbf{M}_{\tilde{A}}^{i}|| \cdot ||\mathbf{M}_{\tilde{B}}^{j}||} \\
&=
 \frac {\mathbf{U}^{T}{\mathbf{M}}_{A}^{i} \cdot \mathbf{V}^{T}{\mathbf{M}}_{B}^{j}}{||\mathbf{U}^{T}{\mathbf{M}}_{A}^{i}|| \cdot ||\mathbf{V}^{T}{\mathbf{M}}_{B}^{j}||}.\label{eq3}
\end{aligned}
\end{equation}

To find the cross-modal anomalies, we need to ensure that for the pair of instances with consistent patterns across different modalities (i.e., $\mathbf{y}_{A}^{i} = \mathbf{y}_{B}^{j}$), their cross-modal similarity in the transformed feature space should be high. To this end, the loss function can be expressed as follows:
\begin{equation}
\begin{aligned}
\min_{\mathbf{U},\mathbf{V}} \sum_ {\{i,j\} \in \mathcal{S}} ^ {} 1-F({\mathbf{M}}_{A}^{i} , {\mathbf{M}}_{B}^{j}), \\
\end{aligned}
\end{equation}
where the set $\mathcal{S}$ contains the instance pairs $\{i,j\}$ whose cross-modal observations are consistent, i.e., $\mathbf{y}_{A}^{i} = \mathbf{y}_{B}^{j}$.

Also, we need to penalize the instance pairs with inconsistent patterns across different modalities. In particular, in the training process, we not only make use of the instance pairs with consistent patterns across different modalities, but also incorporate an additional set of negative samples by negative sampling~\cite{mikolov2013exploiting}. Let $\mathcal{N} = \{(p,q)\}$ be the set of the negative samples where their cross-modal patterns are inconsistent (i.e., $\mathbf{y}_{A}^{p} \neq \mathbf{y}_{B}^{q}$), then to learn the mapping matrices, the loss function for encouraging dissimilarities of negative samples can be presented in the following format:
\begin{equation}
\begin{aligned}
\min_{\mathbf{U},\mathbf{V}}  \sum_{(p,q) \in \mathcal{N}} ^ {} \max \left(0, F({\mathbf{M}}_{A}^{p} , {\mathbf{M}}_{B}^{q}) - \gamma \right).\\
\end{aligned}
\end{equation}
In the above formulation, the hyperparameter $\gamma$, as a margin value, controls the penalty degree of the inconsistent patterns. Specifically, if $F({\mathbf{M}}_{A}^{p},{\mathbf{M}}_{B}^{q})> \gamma$, it implies that the instance pairs with inconsistent patterns across different modalities are regarded as similar and it is necessary to penalize it in the loss function.

By combining Eq. (5) and Eq. (6), it leads to the following objective function for the transformation matrices learning:
\begin{equation}
\begin{aligned}
\min_{\mathbf{U},\mathbf{V}} & \sum_ {\{i,j\} \in \mathcal{S}} ^ {} 1-F({\mathbf{M}}_{A}^{i} , {\mathbf{M}}_{B}^{j})\\
& +  \lambda_{0}\!\!\!\sum_{(p,q) \in \mathcal{N}} ^ {} \max \left(0, F({\mathbf{M}}_{A}^{p} , {\mathbf{M}}_{B}^{q}) - \gamma \right)\\
& +\lambda_{1} (||\mathbf{U}||_{F}^{2}+||\mathbf{V}||_{F}^{2}). \\
\end{aligned}
\end{equation}
The above loss function consists of three terms. The first term is to ``pull'' the projections of different modalities together if their cross-modal patterns (w.r.t. instance class labels) are consistent, while the second term is to ``push'' them further apart otherwise (as shown in Figure 1(b)). The resultant unified feature space is shown in Figure 1(c). The parameter $\lambda_{0}$ controls the influence of negative sampling. The third term is a regularization term, which controls the bias-variance trade-off in order to avoid the over-fitting, and $\lambda_{1}$ is the regularization hyper-parameter. The aforementioned objective function can be effectively solved with coordinate descent methods, in which the variables $\mathbf{U}$ and $\mathbf{V}$ are updated in an alternating way until the objective function converges to a local optimum.

Deep neural networks have been successfully applied as powerful feature extraction models for different types of data with single modality, including text, image and audio data~\cite{ngiam2011multimodal}. The main reason for the success can be attributed to the fact that deep neural networks are capable of learning nonlinear mappings by extracting high-level abstractions from the input raw features. Over the past few decades, many successful deep neural networks such as the deep Boltzmann machines~\cite{salakhutdinov2010efficient}, auto-encoders~\cite{ngiam2011multimodal}, and recurrent neural networks~\cite{graves2013speech,huang2019graph} have been widely applied and achieved the state-of-the-art performance in many learning tasks.

Motivated by the success of deep neural networks, we develop a deep structured framework to characterize the features of different modalities for cross-modal anomaly detection. In particular, we employ a series of nonlinear mapping functions in the deep neural networks to map information of each modality into a consensus feature space, in which the instance pairs with consistent patterns across different modalities are pushed away while the pairs with inconsistent cross-modal patterns are pulled together. In this way, the trained model can help us identify the cross-modal anomalies in the test data. Compared with previously introduced linear projections, the developed method employs deep neural networks to learn nonlinear mappings. Thus it empowers a stronger ability to learn more effective feature representations from the original multi-modal data~\cite{liu2019single}. In addition, the nonlinear mapping functions help to fully capture the nonlinear correlations among different modalities, which is otherwise difficult to characterize with conventional linear projection functions. It should be noted that the previously mentioned linear mapping is mathematically equivalent to imposing a single fully connected layer (without nonlinear activation functions) in the deep neural networks. Specifically, if we replace the linear mapping process (through mapping matrices $\mathbf{U},\mathbf{V}$)  with neural networks, then we have the loss function as:

\begin{equation}
\begin{aligned}
\min_{\mathcal{W}} &  \sum_ {\{i,j\} \in \mathcal{S}} ^ {}  1-F\left(f({\mathbf{M}}_{A}^{i}), g({\mathbf{M}}_{B}^{j})\right) \\
 + & \lambda \sum_{\{p,q\} \in \mathcal{N}} ^ {} \max \left(0, F(f({\mathbf{M}}_{A}^{p}) , {g(\mathbf{M}}_{B}^{q})) - \gamma \right),\\
\end{aligned}\label{eq8}
\end{equation}
where $f(\cdot)$ and $g(\cdot)$ denote neural network-like differentiable functions, $\mathcal{W}$ denotes the set of all parameters of the deep neural network. We determine the architecture of the neural networks based on the data modalities to be handled. Specifically, we extract features from images through a convolutional neural network (CNN), and train a fully connected neural network to process text tags. The parameter $\lambda$ controls the importance of the second term. In the experiments, we empirically set $\lambda$ as 1.

Our goal is to minimize Eq. (\ref{eq8}) through updating all the model parameters $\mathcal{W}$ in the deep neural networks. With an initialization of model parameters, the proposed deep model CMAD is optimized by Adam
, which is an adaptive momentum based gradient descent method. The detailed are given in Algorithm \ref{ag1}. We make use of the dropout~\cite{srivastava2014dropout} in the training process to avoid over-fitting. After the optimization process, the instances with consistent patterns across different modalities will be pulled together within a small distance, while the instances with inconsistent patterns across different modalities will be pushed away from each other. Finally, for example, the distribution of data instances in the projected latent space is shown as Figure 1 (c). Thus, we can use the CMAD to identify cross-modality anomalies by measuring the similarity across different modalities as shown in Algorithm \ref{ag2}. We show how the performance of CMAD varies with different hyperparameter settings of $\gamma$ in Eq. (\ref{eq8}) and the threshold $\epsilon$ in Eq.(\ref{eq1}) in the section Experiment D.

\begin{algorithm}[t]
\caption{The training process of CMAD.}
\label{ag1}
\begin{algorithmic}[1]
\State \textbf{Input:} Initialize a dataset $\mathcal{X}$, which contains multiple modalities as the data source, such as $\mathbf{M}_{A}$, $\mathbf{M}_{B}$
\For{each epoch}
    \For{each mini batch}
        \For{the positive training samples $\{i,j\}$ in $\mathcal{S}$}
              \State $loss+= 1-F({\mathbf{M}}_{A}^{i} , {\mathbf{M}}_{B}^{j})$;
        \EndFor
        \For{the negative training samples $\{p,q\}$ in $\mathcal{N}$ from negative sampling} 
            \State  { $loss+= \max(0, F({\mathbf{M}}_{A}^{p} , {\mathbf{M}}_{B}^{q}) - \gamma)$;}
          \EndFor
    \EndFor
       \State {Update the model parameters with Adam;}
\EndFor
\State \textbf{Output:} The learned model parameters of CMAD.
\end{algorithmic}

\end{algorithm}

\begin{algorithm}[t]
\caption{Anomaly detection with CMAD.}
\label{ag2}
\begin{algorithmic}[1]
\State \textbf{Input:} instances from two modalities, $\mathbf{M}_{A}, \mathbf{M}_{B}$
\State \textbf{Output:} the cross-modal anomalies
\State load the learned model parameters from Algorithm 1;
\For{$i$ $\in$ \{1,2,...$N$\}}
  \If {$F({\mathbf{M}}_{A}^{i} , {\mathbf{M}}_{B}^{i}) - \epsilon < 0$}
  \State the $i^{th}$ instance is a cross-modal anomaly;
  \EndIf
\EndFor
\end{algorithmic}
\end{algorithm}

\section {Experiments}
In this section, we conduct experiments to evaluate the effectiveness of our proposed CMAD framework in detecting cross-modal anomalies. We also perform a case study to visualize the distribution in the latent spaces to illuminate effectiveness of the CMAD. 

\subsection{Datasets}
We use two real-world cross-modal datasets to assess the effectiveness of the proposed framework in cross-modal anomaly detection. The details of the datasets are as follow:

\begin{itemize}
  \item \textbf{MNIST}~\cite{lecun-mnisthandwrittendigit-2010}: In the MNIST dataset, we have 60,000 original images to represent ten different digits of 1$\times$28$\times$28 pixels. After adding the text tag to each image and embed tags through different word embedding methods(Word2Vec~\cite{mikolov2013distributed}, GloVe~\cite{pennington2014glove}) to 100 dimension vectors, we can synthetically generate the training data with two different modalities. The image modality is fed into a CNN model, and the text modality is fed into a fully connected neural network for training.
    \item \textbf{RGB-D}~\cite{lai2011large}: In the RGB-D set, we have two modalities, RGB images and depth images to represent the same objects in the real-world. In our experiments, we include 960  RGB images and 960 corresponding depth images from five different kinds of objects in the RGB-D dataset, including apples, staplers, coffee-mugs, soda-cans, and toothpastes. Both modalities are fed into a CNN model for feature learning.
\end{itemize}

As there is no ground truth of anomalies in these datasets, thus we need to inject anomalies for evaluation. To generate the cross-modal anomalies, we adopt a widely used injection method - negative sampling, to create a number of instance pairs such that their patterns are inconsistent across different modalities. Originally, we only have instances whose patterns are consistent across different modalities (with the same class labels). To introduce the inconsistent instance pairs, we randomly sample a number of $k$ instance pairs $\{p,q\}$ such that their cross-modal patterns are different such that $\mathbf{y}_{A}^{p} \neq \mathbf{y}_{B}^{q}$. To be more specific, in the testing stage, we inject 1,019 inconsistent pairs in the MNIST, and 545 inconsistent pairs in the RGB-D datasets. In the meanwhile, we have the same portion of instances with consistent behavior which have no overlap with the training stage.

\begin{table}[t]
\caption{Hyaperparameters of the deep neural models.}
\begin{center}
\scalebox{1}[1]{
  \begin{tabular}{|c|c|}
  \hline
  \textbf{Dataset}&  Dimension of the representations \\
  & in the hidden layers\\
  \hline
  MNIST Images &784-1440-1280-320-150-50     \\
  MNIST Tags &100-100-50     \\
  RGB-D RGB Images&58575-1440-1280-320-150-50     \\
  RGB-D Depth Images&58575-1440-1280-320-150-50      \\

  \hline
  \end{tabular}}
\label{tab2}
\end{center}
\end{table}

\subsection{Baseline Algorithms}
We introduce several widely used baseline methods to compare the performance of cross-modal anomaly detection on multi-modal data:

\begin{itemize}
  \item  \textbf{CCA \& KCCA}~\cite{thompson2000canonical}. CCA models correlation across multiple modalities and implicitly maps instances into a lower-dimensional space, with the target to find the maximum correlations in the latent space by linear projections. Compared with CCA, KCCA ~\cite{lai2000kernel,akaho2006kernel} uses an alternative projection strategy by nonlinearly mapping the multi-modal data into a consensus feature space with kernel tricks. We maximize the correlations of instances with consistent patterns in the training phase, and identify cross-modal anomalies by their cross-modal correlation in the test phase.
  Both CCA and KCCA are supervised.
  
  \item  \textbf{PLS}~\cite{abdi2003partial}. PLS is also a supervised feature representation learning method based on linear transformation. The major difference between PLS and CCA is that PLS maximizes the covariance between different modalities through linear transformations. We identify the cross-modal anomalies in the same way as CCA and KCCA.

\item  \textbf{HOAD}~\cite{gao2011spectral}. HOAD is a supervised anomaly detection algorithm to find anomalies from the multi-modal data. It first obtains the spectral embeddings from the multi-modal data with an ensemble similarity matrix, and then calculates the anomaly score of each instance based on the cosine distance between different embeddings.

\item \textbf{Embedding Network }~\cite{wang2018learning,wang2016learning}. An supervised representation learning framework based on deep neural networks to extract features from different modalities, such as image and text. We distinguish the cross-modal anomalies by measuring the euclidean distances across different modalities after projecting instances into the latent space.
\end{itemize}

In our experiment, we evaluate the performance of cross-modal anomaly detection using the metrics of precision, recall, and accuracy. 

We use TP, FP, TN, FN as the number of true positives, false positives, true negatives and false negatives, respectively, in the predicted results. Their definitions are listed as follows:

\begin{equation}
\mathrm{precision} = \frac{TP}{TP+FP},
\end{equation}

\begin{equation}
\mathrm{recall} = \frac{TP}{TP+FN},
\end{equation}

\begin{equation}
\mathrm{accuracy} = \frac{TP+FP}{TP+TN+FP+FN}.
\end{equation}

\subsection{Anomaly Detection Performance}
The parameter settings in our deep model CMAD vary among different modalities and datasets. Specifically, we use convolutional neural network (CNN) for the image modality~\cite{zhai2016deep,chang2015heterogeneous}, and fully-connected neural network for the text tag modality. The hyperparameters of the feature dimensionality in each layer are listed in Table 2. Table 3 and Table 4 present the performance of cross-modal anomaly detection using different algorithms, which include the accuracy, precision and recall of the anomaly detection results.

As we can see from the tables, several observations are drawn as below. 

\begin{table}[t]
\caption{Performance on MNIST.}
\begin{center}
\scalebox{1}[1]{
  \begin{tabular}{|c|c|c|c|}
  \hline

  \textbf{Method}& \textbf{Accuracy}&\textbf{Precision} &\textbf{Recall} \\

  \hline
  CCA& 0.6654&0.8911 &0.6191 \\
  Kernel CCA(RBF)&0.6923&0.7062&0.6948 \\
  PLS &0.7235 &0.6636&0.6233\\
  HOAD &0.5118&0.9220&0.5115\\
  Embedding Network&0.9504&0.9873&0.9127\\
  \textbf{CMAD}& \textbf{0.9921} & \textbf{0.9971} & \textbf{0.9875}  \\


  \hline
  \end{tabular}}
\label{tab4}
\end{center}
\end{table}
\begin{table}[t]
\caption{Performance on RGB-D.}
\begin{center}
\scalebox{1}[1]{
  \begin{tabular}{|c|c|c|c|}
  \hline

  \textbf{Method}& \textbf{Accuracy}&\textbf{Precision} &\textbf{Recall} \\

  \hline
  CCA& 0.7682&0.7171 &0.7988 \\
  Kernel CCA(RBF)&0.8456&0.7134&0.9699 \\
  PLS &0.6580 &0.8304&0.7539\\
  HOAD &0.5137&\textbf{0.9345}&0.5124\\
  Embedding Network&0.5138&0.5075&0.9346\\
  \textbf{CMAD}& \textbf{0.9512} & 0.9313 & \textbf{0.9742}  \\


  \hline
  \end{tabular}}
\label{tab4}
\end{center}
\end{table}

First, CMAD outperforms the baseline methods CCA, KCCA, PLS and HOAD in most cases, which validates the importance of applying deep models for cross-modal anomaly detection by extracting more effective feature representations from the original multi-modal data.

Second, we compare CMAD against the Embedding Network method. Similar to our proposed CMAD, the Embedding Network is also a deep structure based approach. Based on the experimental results, both of the deep structured methods outperform other linear transformation based methods. These observations validate the limitation of the linear projection based methods, i.e., the linear transformation, as they cannot fully capture nonlinear correlations among different modalities for cross-modal anomaly detection.

There are two major differences between the proposed CMAD and the Embedding Network method. First, Embedding Network is developed to handle the text and image data only, while CMAD is able to handle different types of data with appropriate deep neural network designs. Second, the Embedding Network narrows the Euclidean distance of the pair of instances with consistent information across different modalities, whereas in CMAD, we not only pull samples with similar patterns across different modalities close to each other, but also push samples with dissimilar patterns across different modalities far away from each other. Thus, CMAD can learn better representations from the multi-modal data.

\begin{figure}[t]
\centerline{\includegraphics[width = .5\textwidth]{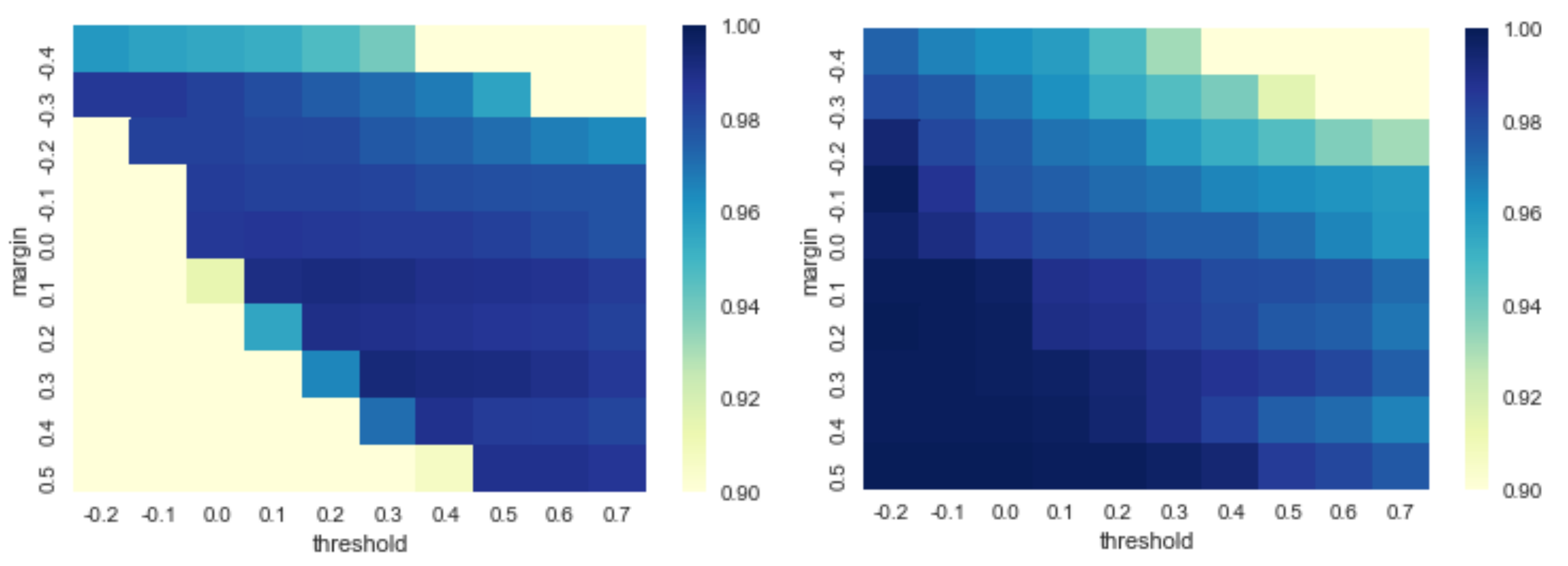}}
\vspace{3mm}
\caption{Precision (left) and recall (Right) in the MNIST dataset with different hyperparameter settings. The figures show how precision and recall scores vary as we change the values of margin $\gamma$ and threshold $\epsilon$.}

\label{fig2}
\end{figure}

\begin{figure}[t]
\centerline{\includegraphics[width = .45\textwidth]{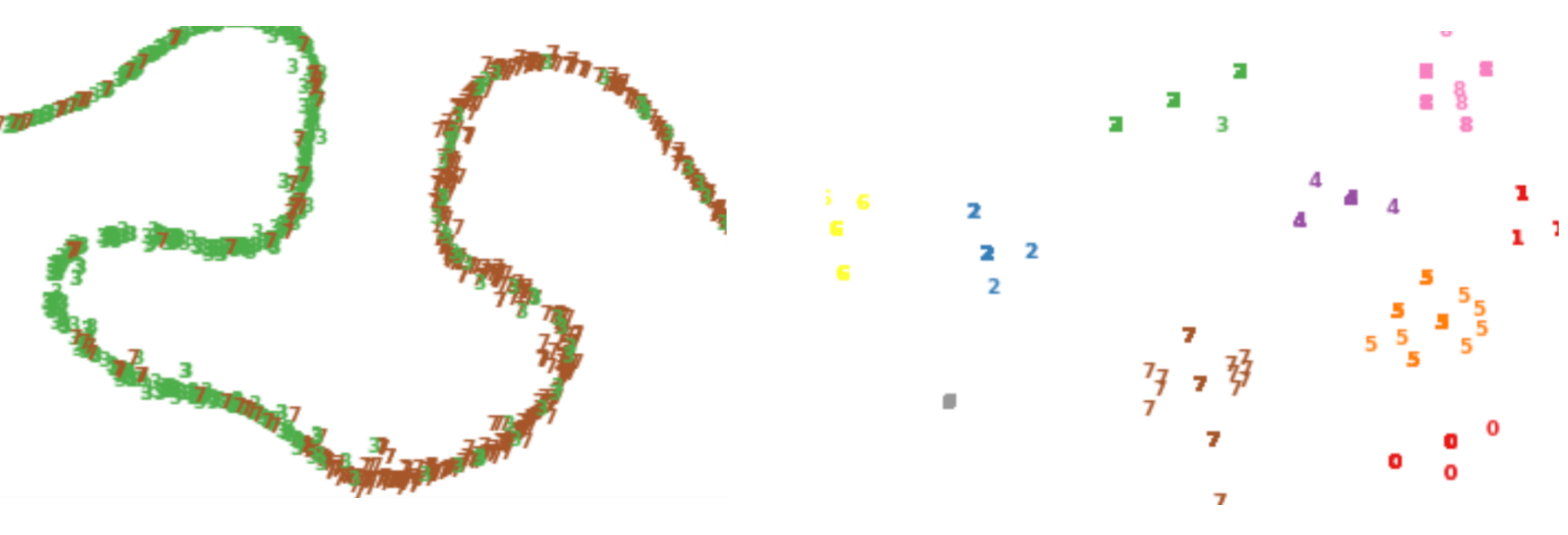}}
\vspace{3mm}
\caption{Left: the distribution of images in the projected space which only contains images  of digits ``3" and ``7" (green denotes ``3", and brown denotes ``7"). Right: the distribution of tags in the projected feature space which contains digits from zero to nine, each color denotes one digit.}

\label{fig8}
\end{figure}

\subsection{Hyperparameter Settings}
There are two hyperparameters to be tuned in our method, i.e., the margin $\gamma$ in Eq. (\ref{eq8}) and the threshold $\epsilon$ in Eq.(\ref{eq1}). In Figure \ref{fig2}, we show how the performance of CMAD varies with different hyperparameter settings. In particular, we tune the values of these hyperparameters within the range of \{-0.4, -0.3, ... ,0,5\} and \{-0.2, -0.1, ... ,0,7\}, respectively. In general, the algorithm performs better w.r.t. precision when the threshold $\epsilon$ is set to be small and the margin $\gamma$ is set to be large. The accuracy achieves the best value when $\epsilon$ is 0.3 and $\gamma$ equals to 0.3. In the meanwhile, we can find that the proposed CMAD could achieve a good performance (over 92\%) in a wide range of hyperparameter settings, the loose condition give us more opportunity to find anomalies in the real-world applications.

\begin{figure}[t]
\centerline{\includegraphics[width = .3\textwidth]{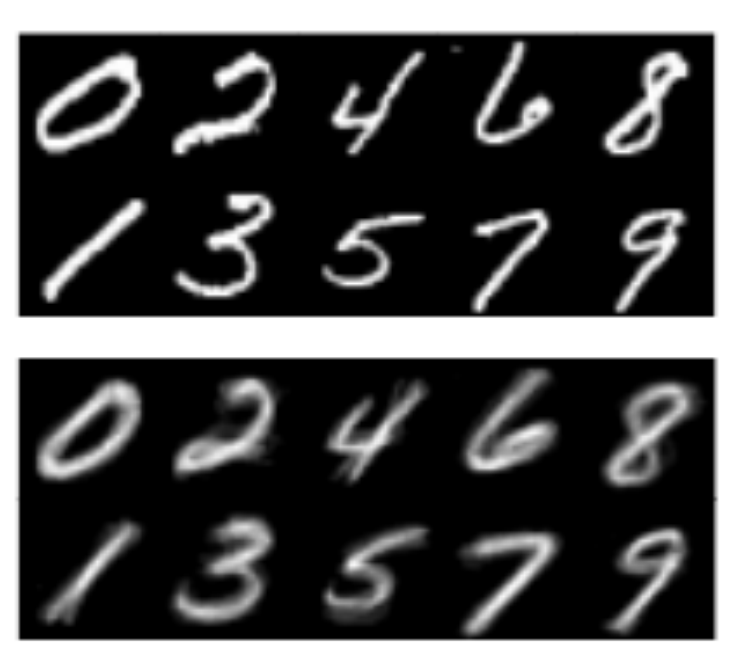}}
\vspace{3mm}
\caption{Reconstruction result in MNIST. According to finding the closest instances from other modalities
, we could demonstrate that, the distances between the same objects from different modalities are closer than those denoting different objects in the latent space. The upper part are the selected images as the ground truth, and the lower part are reconstructed images through finding the closest neighbors which originally distributed in the text tag space.}
\label{fig12}
\end{figure}

\subsection{Case Studies}
Finally, we conduct two case studies to visualize the data distribution in the projected consensus feature space.  Both of the cases are based on the MNIST dataset. The case studies are separated into two parts. The first part is to represent the data distribution of each individual modality in the latent space, and to demonstrate the effectiveness of the ``pull'' and ``push'' actions in our developed method. In the second part, we demonstrate the representation ability of CMAD across different modalities.  

In the first case, in order to show that CMAD is capable to separate different instances in the projected consensus feature space, we visualize the data distribution in individual modalities after the training stage. To represent the data distribution, we first map the original instances into the projected feature space. Then we further reduce the dimensionality of the projected feature space to a two-dimensional visible feature space by PCA. As shown in Figure 3, in each modality, the data instances which have similar patterns are separated from each other. In Figure 3(left), we can find that when the images with labels ``three'' and ``seven'' are taken as the test data, most of the images with label ``three'' (green in the figure) are separated from with the images with label ``seven'' (brown in the figure) in the projected two-dimensional space. In Figure 3 (right), we use the text tag as another data modality. We can find that the distributions of different tags could also be separated in the projected consensus feature space.

The second case is to demonstrate the representation ability of the developed method across different modalities. In other words, our target is to measure whether the distance between observations with similar modalities is closer than the distance between observations with dissimilar modalities after feature mapping. We extract one specific instance from one modality $\mathbf{M}_{A}$ in the projected consensus feature space and find its nearest (based on cosine distance) instances from the other modality $\mathbf{M}_{B}$. 

As shown in Eq. (\ref{eq9}), given an observation $\mathbf{M}_{A}^{i}$ where modality $A$ denotes text data, we extract a set of semantically similar image observations from $\mathbf{M}_{B}$, and aggregate these observations to generate a new image:

\begin{small}
\begin{equation}
\begin{aligned}
  \tiny{\sum_{ \forall M_{B}^{j} \in \mathbf{M_{B}}}  \mathbb{I}  \left({\frac {f({\mathbf{M}}_{A}^{i}) \cdot g({\mathbf{M}}_{B}^{j})}{||f({\mathbf{M}}_{A}^{i})|| \cdot || g({\mathbf{M}}_{B}^{j})||}  - \epsilon > 0 } \right) \cdot {\mathbf{M}}_{B}^{j}}.
\end{aligned}\label{eq9}
\end{equation}
\end{small}

After normalizing the generated image from Eq.(\ref{eq9}) 
in the range of [0, 255] for visualization in the gray scale, we generate a new image which can represent the estimation of the distribution close to the given text $\mathbf{M}_{A}^{i}$. 

We extract instances from the text modality to find the instances from the image modality with the closest cosine distances in the latent space, and reconstruct images from the text features. After normalizing the nearest images, we represent them as new, reconstructed images in the gray scale. The reconstructed results are reported in Figure 4. The upper part contains the ground-truth images, and the lower part denotes to the reconstructed results. Comparing the reconstructed results with the original images, we can intuitively find that our algorithm is effective in pulling images with the consistent instances across different modalities together, and pushing those with inconsistent instances apart away.

\section{Conclusions and Future Work}
In this paper, we propose an novel cross-modal anomaly detection approach CMAD based on deep neural networks. The proposed CMAD framework is able to identify inconsistent patterns or behaviors of instances across different modalities by capturing their nonlinear correlations in the learned consensus latent feature space. Firstly, we train deep structured model to represent features from different modalities, and then project the features into a consensus latent feature space. Secondly, we ``pull'' the projections of a pair of instances from different modalities together if their cross-modal patterns are consistent, while ``push'' them further apart otherwise.  Finally, we distinguish the cross-modality anomalies by measuring the distances across different modalities. We demonstrate the effectiveness of the developed approach on different multi-modal datasets and the experimental results show that CMAD achieves better detection performance than other prevalent anomaly detection methods on multi-modal data. 

There are a number of directions that can be explored for future research, including: (1) extend the current cross-modal anomaly detection framework to other complex data, such as sequential data and networked data; (2) develop a principled way to handle the dynamic data that often arrives at a fast pace, since in many real-world applications the detection of fraudulent or malicious behaviors in real-time is often desired.

\section*{Acknowledgement}
The work is, in part, supported by National Science Foundation (\#IIS-1657196, \#IIS-1750074, \#CNS-1816497).
\newpage

\bibliographystyle{IEEEtran}
\bibliography{reference}
\end{document}